\documentclass{neus2025}

\title[LANTERN: LLM-Augmented Neurosymbolic Transfer]{LANTERN: LLM-Augmented Neurosymbolic Transfer with Experience-Gated Reasoning Networks}

\usepackage{times}
\usepackage{soul}
\usepackage{url}

\usepackage{graphicx}
\usepackage{amsmath}
\usepackage{booktabs}
\usepackage{algorithm}
\usepackage{algpseudocode}
\usepackage{paralist}

\author{%
 \Name{Mahyar Alinejad}$^{1}$ \Email{mahyar.alinejad@ucf.edu}\\
 \Name{Yue Wang}$^{1,2}$ 
 \Email{yue.wang@ucf.edu}\\
 \Name{Amrit Singh Bedi}$^{2}$ 
 \Email{amritbedi@ucf.edu}\\
 \Name{George Atia}$^{1,2}$ \Email{george.atia@ucf.edu}\\
 \addr $^{1}$Department of Electrical and Computer Engineering, University of Central Florida, Orlando, Florida, USA \\
 \addr $^{2}$Department of Computer Science, University of Central Florida, Orlando, Florida, USA
}

\begin{document}

\maketitle

\begin{abstract}
Transfer learning in reinforcement learning (RL) seeks to accelerate learning in new tasks by leveraging knowledge from related sources. Existing neurosymbolic transfer methods, however, typically rely on manually specified task automata, assume a single source task, and use fixed knowledge-integration mechanisms that cannot adapt to varying source relevance.
We propose LANTERN, a unified framework for multi-source neurosymbolic transfer that addresses these limitations through three components: (i) deterministic finite automata generated from natural language task descriptions using large language models, (ii) semantic embedding-based aggregation of multiple source policies weighted by cross-task similarity, and (iii) adaptive teacher-student gating based on temporal-difference error and semantic uncertainty.
Across domains spanning resource management, navigation, and control, LANTERN achieves 40–60\% improvements in sample efficiency over existing baselines while remaining robust to poorly aligned sources. These results demonstrate that multi-source, adaptively weighted neurosymbolic transfer can improve scalability and robustness in symbolic RL settings.
\end{abstract}


\section{Introduction}

Reinforcement learning (RL) has achieved strong empirical performance in game playing~\citep{Mnih2015HumanLevel,Silver2016Go}, robotics~\citep{Kober2013Reinforcement,levine2016end}, and autonomous systems~\citep{Kiran2021Deep}. However, effective policy learning often requires extensive interaction, limiting applicability when data collection is costly or unsafe~\citep{dulac2019challenges}. Transfer learning mitigates this by leveraging knowledge from related tasks~\citep{Taylor2009Transfer,zhu2023transfer}, yet existing approaches remain challenged by structured, long-horizon objectives that are naturally non-Markovian.

Neurosymbolic RL integrates symbolic task representations (such as deterministic finite automata (DFAs) or reward machines~\citep{ToroIcarte2018UsingRL,Icarte2022Reward}) into learning. By encoding temporal structure through product MDP constructions~\citep{Bacchus1997NMRDP}, these methods improve sample efficiency for complex tasks. Despite these advances, several limitations remain.
\textbf{1) Manual specification:} Most methods assume expert-provided DFAs or temporal logic formulas~\citep{Littman2017Environment,Camacho2019LTL,Hahn2019Omega}. Grammatical inference approaches can recover automata from demonstrations~\citep{Angluin1987Learning,Oncina1992,Alinejad2024Hybrid,Alinejad2026Dynamic}, but they require structured trajectory data and are difficult to apply in sparse or exploratory RL settings.
\textbf{2) Single-source transfer:} Automaton distillation transfers symbolic guidance via DFA transitions~\citep{Singireddy2023AutomatonDistillation,Alinejad2025NEUS}, while policy distillation provides action-level knowledge~\citep{Rusu2015PolicyDistillation}. CADENT combines both using experience-based gating~\citep{Alinejad2026Hybrid}. However, these approaches rely on a single source task, which can limit effectiveness when source-target alignment varies.
\textbf{3) Fixed integration mechanisms:} Existing methods typically employ predetermined weighting schemes (e.g., exponential decay~\citep{Singireddy2023AutomatonDistillation}) or static hyperparameters, limiting adaptability when source relevance changes across states or over time.

\noindent\textbf{Key insight and technical novelty.}
We consider neurosymbolic transfer in a setting where multiple source tasks may have partially related but distinct goals from the target. In this regime, transfer cannot rely on direct reuse of a single source policy or automaton; instead, it requires semantic alignment and aggregation of structured knowledge across heterogeneous tasks.

To address this setting, we introduce \textbf{LANTERN} (\textbf{L}LM-\textbf{A}ugmented \textbf{N}eurosymbolic \textbf{T}ransfer with \textbf{E}xperience-gated \textbf{R}easoning \textbf{N}etworks). LANTERN integrates three components. First, DFAs are generated from natural language task descriptions using large language models (LLMs), eliminating manual specification. Second, we construct a shared embedding space over automaton state descriptions, enabling aggregation of partial knowledge from multiple source tasks with heterogeneous goals. Third, we introduce a dual-volatility gating mechanism that combines semantic alignment (measured via embedding similarity) with experience-based reliability (measured via TD error), allowing adaptive weighting of teacher influence during learning.


\noindent\textbf{Contributions.}
Our contributions are threefold: \\
\noindent 1) We formulate multi-source neurosymbolic transfer in a setting where source tasks may have heterogeneous goals, requiring semantic aggregation rather than direct reuse of a single source policy or automaton.

\noindent 2) We develop LANTERN, which integrates LLM-based automaton generation, semantic multi-source aggregation, and adaptive trust gating within a single neurosymbolic transfer architecture. 

\noindent 3) Across diverse domains, we demonstrate 40--60\% improvements in sample efficiency over single-source and static-integration baselines, while maintaining robustness to poorly aligned sources.

\subsection{Related Work}

\noindent\textbf{Transfer learning in RL.}
Classical transfer methods include value function reuse~\citep{Taylor2007Cross}, policy distillation~\citep{Rusu2015PolicyDistillation,Czarnecki2019Distilling}, and successor features~\citep{Barreto2017Successo,Barreto2020Fast}. Meta-learning~\citep{Finn2017Model,rakelly2019efficient} and multi-task learning~\citep{Parisotto2015ActorMimic,teh2017distral} share representations across related tasks. These approaches typically assume Markovian reward structures and do not explicitly model temporal logic or automaton-based task decomposition.

\noindent\textbf{Neurosymbolic RL.}
To address non-Markovian objectives, reward machines~\citep{ToroIcarte2018UsingRL,Icarte2022Reward} and temporal logic specifications~\citep{Littman2017Environment,Camacho2019LTL,Hahn2019Omega} encode structured task progression via product MDP constructions. Extensions integrate automata with deep RL~\citep{Hasanbeig2020Deep,DeGiacomo2019Shielding} or infer specifications from demonstrations~\citep{VazquezChanlatte2018LearningSpecs}. However, these works focus primarily on single-task learning rather than transfer across heterogeneous tasks.

\noindent\textbf{Automaton-based transfer.}
Recent work leverages automaton structure for transfer. Automaton distillation~\citep{Singireddy2023AutomatonDistillation} transfers high-level task decomposition through DFA-guided Q-value aggregation. Bidirectional transfer frameworks~\citep{Alinejad2025NEUS} enable mutual knowledge exchange, while CADENT~\citep{Alinejad2026Hybrid} combines strategic automaton guidance with tactical policy distillation using experience-based gating. ARM-FM~\citep{Creus2024ARMFM} employs LLM-generated reward machines for transfer. These approaches, however, rely on single-source settings and fixed or experience-only integration mechanisms.

\noindent\textbf{LLMs in RL.}
LLMs have been used to provide planning guidance~\citep{Jiang2019Language}, programmatic policy representations~\citep{Verma2018Programmatically,andreasmodular}, and zero-shot generalization signals~\citep{Oh2017Zero}. In contrast to approaches that use language primarily for prompting or reward shaping, LANTERN generates formal DFAs compatible with product MDP constructions and integrates them into a multi-source neurosymbolic transfer framework.

The remainder of this paper is structured as follows: Section~\ref{sec:background} provides necessary background on product MDPs and transfer learning. Section~\ref{sec:method} details the LANTERN framework. Section~\ref{sec:experiments} reports experimental results across four domains. Section~\ref{sec:conclusion} concludes with future directions. 

\section{Background}
\label{sec:background}

\subsection{Markov Decision Processes and Q-Learning}

A Markov Decision Process (MDP)~\citep{SuttonBarto} is a tuple $\mathcal{M} = \langle \mathcal{S}, \mathcal{A}, \mathcal{T}, \mathcal{R}, \gamma \rangle$, where $\mathcal{S}$ is the state space, $\mathcal{A}$ is the action space, $\mathcal{T}: \mathcal{S} \times \mathcal{A} \times \mathcal{S} \to [0,1]$ is the transition function, $\mathcal{R}: \mathcal{S} \times \mathcal{A} \to \mathbb{R}$ is the reward function, and $\gamma \in [0,1)$ is the discount factor. A policy $\pi: \mathcal{S} \to \Delta(\mathcal{A})$ maps states to action distributions, where $\Delta(\mathcal{A})$ is the probability simplex on $\mathcal{A}$.

The goal is to find $\pi^* = \arg\max_\pi \mathbb{E}_{\pi} [ \sum_{t=0}^{\infty} \gamma^t \mathcal{R}(s_t, a_t) \mid s_0 ]$. The optimal action-value function $Q^*(s,a) = \max_\pi Q^\pi(s,a)$ satisfies:
\begin{equation}
Q^*(s,a) = \mathbb{E}_{s'} \left[ \mathcal{R}(s,a) + \gamma \max_{a'} Q^*(s',a') \right].
\end{equation}

\noindent Q-learning~\citep{Watkins1992QLearning} iteratively estimates $Q^*$ via:
\begin{equation}
Q(s_t, a_t) \leftarrow Q(s_t, a_t) + \alpha \left[ \mathcal{R}(s_t, a_t) + \gamma \max_{a'} Q(s_{t+1}, a') - Q(s_t, a_t) \right],
\end{equation}
where $\alpha \in (0,1]$ is the learning rate.

\subsection{Product MDPs for Non-Markovian Objectives}

Many tasks involve non-Markovian objectives depending on state history~\citep{Bacchus1997NMRDP}. A DFA $\mathcal{D} = \langle \Omega, \Sigma, \delta, \omega_0, F \rangle$ specifies such tasks, where $\Omega$ is the set of automaton states, $\Sigma$ is the set of labels, $\delta: \Omega \times \Sigma \to \Omega$ is the transition function, $\omega_0$ is the initial state, and $F \subseteq \Omega$ are accepting states. A labeling function $L: \mathcal{S} \to \Sigma$ maps MDP states to labels.

The product MDP $\mathcal{M} \times \mathcal{D} = \langle \mathcal{S} \times \Omega, \mathcal{A}, \mathcal{T}', \mathcal{R}', \gamma \rangle$ has state space $\mathcal{S} \times \Omega$ with $(s, \omega)$ representing the agent at MDP state $s$ with the automaton in state $\omega$. The transition function is $\mathcal{T}'((s,\omega), a, (s',\omega')) = \mathcal{T}(s,a,s')$ if $\omega' = \delta(\omega, L(s'))$, and $0$ otherwise. The reward function $\mathcal{R}'((s,\omega), a)$ is designed based on automaton progress, typically providing sparse rewards when reaching accepting states ($\omega' \in F$) or incremental rewards when making automaton transitions ($\omega \neq \omega'$). This transforms non-Markovian objectives into standard MDP learning by tracking task progress through $\omega$~\citep{ToroIcarte2018UsingRL,Icarte2022Reward}.



\subsection{Transfer Learning and Neurosymbolic Methods}

Transfer learning accelerates target task learning by leveraging source knowledge~\citep{Taylor2009Transfer}. \textbf{Policy distillation}~\citep{Rusu2015PolicyDistillation} trains student $\pi^{\text{student}}$ to mimic teacher $\pi^{\text{teacher}}$ by minimizing $D_{\text{KL}}(\pi^{\text{teacher}}(\cdot|s) \| \pi^{\text{student}}(\cdot|s))$.

\textbf{Automaton distillation}~\citep{Singireddy2023AutomatonDistillation, Alinejad2025NEUS} transfers strategic knowledge via DFA transitions. Given a teacher Q-function, $Q^{\text{teacher}}$, learned on a source product MDP, the method computes aggregated Q-values for each automaton transition $(\omega, \omega')$:
\begin{equation}
Q_{\text{AD}}(\omega, \omega') = \frac{1}{|\mathcal{S}_{\omega \to \omega'}|} \sum_{(s,a) \in \mathcal{S}_{\omega \to \omega'}} Q^{\text{teacher}}((s,\omega), a),
\label{eq:automaton_distillation}
\end{equation}
where $\mathcal{S}_{\omega \to \omega'} = \{(s,a) : \delta(\omega, L(s')) = \omega', s' \sim \mathcal{T}(s,a,\cdot)\}$ is the set of state-action pairs that trigger the automaton transition from $\omega$ to $\omega'$, and $Q^{\text{teacher}}((s,\omega), a)$ is the teacher's learned action-value function on the product MDP. During target task learning, the student receives additional reward $\lambda_{\text{AD}} \cdot Q_{\text{AD}}(\omega, \omega')$ when making automaton transitions.

\textbf{CADENT} combines strategic and tactical guidance with experience-based gating~\citep{Alinejad2026Hybrid}. It tracks temporal-difference (TD) error volatility (a measure of learning instability) for each state-action pair:
$V_t(s,a) \leftarrow (1-\eta) V_{t-1}(s,a) + \eta |\delta_t(s,a)|,$ where $\delta_t(s,a) = r_t + \gamma \max_{a'} Q_t(s',a') - Q_t(s,a)$ is the TD error at time $t$, and $\eta \in (0,1)$ is a smoothing parameter. A trust gate measuring confidence in the student’s own estimate is computed as $
\tau(s,a) = \sigma\!\left(-k\big(V(s,a) - \theta\big)\right),$
where $\sigma(x) = 1/(1+e^{-x})$ is the sigmoid function, $k > 0$ controls gate sharpness, and $\theta \in (0,1)$ is a threshold. The Q-update balances student learning and teacher guidance:
\begin{equation}
\Delta Q(s,a) = \alpha \left[ 
\tau(s,a)\, \delta_t(s,a) 
+ (1-\tau(s,a))\, G_{\text{teacher}}(s,a) 
\right],
\label{eq:cadent_update}
\end{equation}
where
\[
G_{\text{teacher}}(s,a)
=
\lambda_{\text{AD}} r_{\text{AD}}(\omega,\omega')
+
\lambda_{\text{PD}}
\big(\pi^{\text{teacher}}(a|s) - \pi^{\text{student}}(a|s)\big),
\]
combines strategic guidance (intrinsic reward $r_{\text{AD}}$ for automaton transitions $\omega \to \omega'$) and tactical guidance (policy discrepancy), weighted by $\lambda_{\text{AD}}, \lambda_{\text{PD}} \ge 0$.


\textbf{ARM-FM}~\citep{Creus2024ARMFM} generates reward machines via LLMs with single-source embedding transfer. However, all existing methods use single sources and lack graceful degradation under misalignment.

\section{Formulation and Proposed LANTERN Framework}
\label{sec:method}

\subsection{Problem Formulation}



Consider $K$ source tasks, each modeled as a product MDP $\mathcal{M}_k^{\text{src}} \times \mathcal{D}_k^{\text{src}}$,  where the automaton 
$\mathcal{D}_k^{\text{src}} = \langle \Omega_k^{\text{src}}, \Sigma_k, \delta_k^{\text{src}}, \omega_{0,k}^{\text{src}}, F_k^{\text{src}} \rangle$  encodes the task structure. For each source task $k$, we assume access to: (i) a learned Q-function $Q_k^{\text{teacher}}$ defined on the product MDP, (ii) distilled strategic knowledge $Q_{k,\text{AD}}$ (Eq.~\ref{eq:automaton_distillation}), (iii) a distilled tactical policy $\pi_k^{\text{teacher}}$, and (iv) semantic descriptions $\mathrm{desc}_k : \Omega_k^{\text{src}} \to \mathcal{V}$ mapping each automaton state to a natural language description in vocabulary space $\mathcal{V}$.

The target task is specified only by a natural language description $\mathcal{T}_{\text{desc}} \in \mathcal{V}$  and a base MDP $\mathcal{M}^{\text{tgt}}$, without a manually provided automaton. Our objective is to learn a target policy $\pi^{\text{tgt}}$ that maximizes expected return on the induced product MDP while leveraging the multi-source knowledge set
$
\mathcal{K} = \{ (Q_k^{\text{teacher}}, Q_{k,\text{AD}}, \pi_k^{\text{teacher}}, \mathrm{desc}_k) \}_{k=1}^K,
$
under the setting where source tasks may have heterogeneous goals and distinct automaton structures. The automaton state spaces $\Omega_k^{\text{src}}$ are not aligned across tasks, and source goals may differ from the target goal, precluding direct reuse of symbolic states or value functions.

\subsection{Phase 1: LLM-Enhanced Automaton Generation}

Given a natural language task description $\mathcal{T}_{\text{desc}} \in \mathcal{V}$, we use a LLM $\mathcal{L}$ to generate a target DFA
$\mathcal{D}^{\text{tgt}} = \langle \Omega^{\text{tgt}}, \Sigma^{\text{tgt}}, \delta^{\text{tgt}}, \omega_0^{\text{tgt}}, F^{\text{tgt}} \rangle,$ together with semantic state descriptions $\mathrm{desc}^{\text{tgt}} : \Omega^{\text{tgt}} \to \mathcal{V}$ that assign each automaton state a natural language description.

\paragraph{Prompt construction.}
We design a structured prompt $\mathcal{P}(\mathcal{T}_{\text{desc}})$ that instructs $\mathcal{L}$ to: 
(i) extract key subgoals and temporal dependencies from $\mathcal{T}_{\text{desc}}$, 
(ii) define automaton states $\Omega^{\text{tgt}}$ representing task-progress milestones, 
(iii) specify a deterministic transition function $\delta^{\text{tgt}} : \Omega^{\text{tgt}} \times \Sigma^{\text{tgt}} \to \Omega^{\text{tgt}}$, 
(iv) designate the initial state $\omega_0^{\text{tgt}}$ and accepting states $F^{\text{tgt}}$, and 
(v) provide a semantic description $\mathrm{desc}^{\text{tgt}}(\omega)$ for each state $\omega \in \Omega^{\text{tgt}}$.




\paragraph{Example.}
Given $\mathcal{T}_{\text{desc}} =$ 
``Navigate dungeon to collect key and shield, then open chest for sword, finally defeat dragon,'' 
the LLM generates a DFA with:

\begin{itemize}
\item States: $\Omega^{\text{tgt}} = \{\omega_0, \omega_1, \omega_2, \omega_3, \omega_4\}$

\item Descriptions:
$\mathrm{desc}^{\text{tgt}}(\omega_0) =$ ``start mission'', 
$\mathrm{desc}^{\text{tgt}}(\omega_1) =$ ``collect key'', 
$\mathrm{desc}^{\text{tgt}}(\omega_2) =$ ``collect shield'', 
$\mathrm{desc}^{\text{tgt}}(\omega_3) =$ ``obtain sword from chest'', 
$\mathrm{desc}^{\text{tgt}}(\omega_4) =$ ``defeat dragon (goal)''

\item Transitions:
$\omega_0 \xrightarrow{\text{key}} \omega_1 
\xrightarrow{\text{shield}} \omega_2 
\xrightarrow{\text{sword}} \omega_3 
\xrightarrow{\text{dragon}} \omega_4$.
\end{itemize}


\paragraph{Product MDP construction.}
A labeling function $L^{\text{tgt}} : \mathcal{S}^{\text{tgt}} \to \Sigma^{\text{tgt}}$ maps environment states to automaton symbols based on observable conditions (e.g., item collection or goal completion events). Given the base MDP $\mathcal{M}^{\text{tgt}}$ and the LLM-generated DFA $\mathcal{D}^{\text{tgt}}$, we construct the product MDP $\mathcal{M}^{\text{tgt}} \times \mathcal{D}^{\text{tgt}}$ with augmented state space $\mathcal{S}^{\text{tgt}} \times \Omega^{\text{tgt}}$, following the standard construction described in Section~\ref{sec:background}. While prior work has used LLMs to generate reward machines or automata~\citep{Creus2024ARMFM}, our use of semantic descriptions extends beyond specification generation. The descriptions $\mathrm{desc}^{\text{tgt}}$ define a semantic representation of automaton states that will later support multi-source knowledge aggregation and adaptive teacher-student gating within the LANTERN framework.

\subsection{Phase 2: Semantic Embedding and Neighborhood Construction}

To enable transfer across heterogeneous source tasks, we construct a shared semantic embedding space over automaton state descriptions. Unlike single-source approaches~\citep{Creus2024ARMFM,Alinejad2026Hybrid}, this embedding allows alignment and aggregation of symbolic states originating from different task goals.

For each automaton state $\omega$ with description $\mathrm{desc}(\omega)$, we compute an embedding
\begin{equation}
\phi(\omega) = \mathcal{E}(\mathrm{desc}(\omega)) \in \mathbb{R}^d,
\end{equation}
where $\mathcal{E}: \mathcal{V} \to \mathbb{R}^d$ is a fixed text-embedding model (e.g., sentence-BERT), and $d$ is the embedding dimension. The same embedding function is used for both source and target automata, yielding a shared semantic space.

\paragraph{Cross-task similarity.}
Given a target state $\omega^{\text{tgt}} \in \Omega^{\text{tgt}}$ and a source state $\omega_k^{\text{src}} \in \Omega_k^{\text{src}}$, we define semantic similarity via cosine similarity
\begin{equation}
\mathrm{sim}(\omega^{\text{tgt}}, \omega_k^{\text{src}})
=
\frac{\phi(\omega^{\text{tgt}})^\top \phi(\omega_k^{\text{src}})}
{\|\phi(\omega^{\text{tgt}})\| \, \|\phi(\omega_k^{\text{src}})\|}.
\label{eq:similarity}
\end{equation}
High similarity indicates alignment in task-progress semantics, even when tasks differ in domain.

\paragraph{Semantic neighborhoods.}
For each target state $\omega^{\text{tgt}} \in \Omega^{\text{tgt}}$, we consider all source automaton states across the $K$ tasks and compute their semantic similarity to $\omega^{\text{tgt}}$. The semantic neighborhood $\mathcal{N}_M(\omega^{\text{tgt}})$ is defined as the set of the $M$ source states with the largest similarity values:
\begin{equation}
\mathcal{N}_M(\omega^{\text{tgt}})
=
\left\{
(\omega_k^{\text{src}}, k)
:
\omega_k^{\text{src}} \in \Omega_k^{\text{src}},
\ \text{ranked among the top-$M$ by }
\mathrm{sim}(\omega^{\text{tgt}}, \omega_k^{\text{src}})
\right\}.
\label{eq:neighborhood}
\end{equation}

For each $(\omega_k^{\text{src}}, k) \in \mathcal{N}_M(\omega^{\text{tgt}})$, we define normalized aggregation weights
\begin{equation}
w(\omega^{\text{tgt}}, \omega_k^{\text{src}})
=
\frac{\max\{\mathrm{sim}(\omega^{\text{tgt}}, \omega_k^{\text{src}}), 0\}}
{\sum_{(\omega_j^{\text{src}}, j) \in \mathcal{N}_M(\omega^{\text{tgt}})}
\max\{\mathrm{sim}(\omega^{\text{tgt}}, \omega_j^{\text{src}}), 0\}},
\label{eq:weights}
\end{equation}
so that more semantically aligned states contribute more strongly to subsequent guidance.

\subsection{Phase 3: Multi-Source Knowledge Aggregation}

LANTERN aggregates both strategic (automaton-level) and tactical (policy-level) guidance from the semantic neighborhood of each target automaton state.

\paragraph{Strategic guidance aggregation.}
For a target automaton state $\omega^{\text{tgt}} \in \Omega^{\text{tgt}}$, we aggregate strategic guidance from semantically aligned source states. Let $Q_{k,\text{AD}}(\omega_k^{\text{src}}, \omega_k'^{\text{src}})$ denote the automaton-distilled Q-value in source task $k$ for the transition $\omega_k^{\text{src}} \to \omega_k'^{\text{src}}$. We define the aggregated strategic value as
\begin{equation}
Q_{\text{AD}}^{\text{agg}}(\omega^{\text{tgt}})
=
\sum_{(\omega_k^{\text{src}}, k) \in \mathcal{N}_M(\omega^{\text{tgt}})}
w(\omega^{\text{tgt}}, \omega_k^{\text{src}})
\,
Q_{k,\text{AD}}(\omega_k^{\text{src}}),
\label{eq:strategic_agg}
\end{equation}
where $Q_{k,\text{AD}}(\omega_k^{\text{src}})$ summarizes the strategic value of progressing from $\omega_k^{\text{src}}$ in source $k$ (e.g., expected intrinsic reward over outgoing transitions). This yields a convex combination of high-level task progression signals across sources.

\paragraph{Tactical guidance aggregation.}
At the action level, we aggregate teacher policies from aligned source states. Let $\pi_k^{\text{teacher}}(a \mid s_k, \omega_k^{\text{src}})$ denote the teacher policy in source $k$ defined over product states. For a target state $(s, \omega^{\text{tgt}})$, we define
\begin{equation}
\pi_{\text{teacher}}^{\text{agg}}(a \mid s, \omega^{\text{tgt}})
=
\sum_{(\omega_k^{\text{src}}, k) \in \mathcal{N}_M(\omega^{\text{tgt}})}
w(\omega^{\text{tgt}}, \omega_k^{\text{src}})
\,
\pi_k^{\text{teacher}}(a \mid s_k, \omega_k^{\text{src}}),
\label{eq:tactical_agg}
\end{equation}
where $s_k$ denotes the mapped source-state context corresponding to the target state $s$. 

\subsection{Phase 4: Dual-Volatility Experience Gating}

LANTERN combines experience-based and semantic uncertainty to adaptively balance student and teacher updates.

\paragraph{Experience volatility.}
We track TD-error volatility:
\begin{equation}
V_t^{\text{exp}}(s,a)
\leftarrow
(1-\eta) V_{t-1}^{\text{exp}}(s,a)
+ \eta |\delta_t(s,a)|,
\label{eq:exp_volatility}
\end{equation}
where $\eta \in (0,1)$. High volatility indicates unstable learning (favoring teacher guidance), while low volatility indicates convergence.

\paragraph{Semantic volatility.}
We define
\begin{equation}
V^{\text{sem}}(\omega^{\text{tgt}})
=
1 -
\max_{k,\;\omega_k^{\text{src}} \in \Omega_k^{\text{src}}}
\mathrm{sim}(\omega^{\text{tgt}}, \omega_k^{\text{src}}),
\label{eq:sem_volatility}
\end{equation}
so small values correspond to well-aligned source states and large values to misalignment.

\paragraph{Composite trust gate.}
We convert both volatility measures into trust coefficients as
\begin{align}
\tau_{\text{exp}}(s,a)
&=
\sigma\!\left(-k_{\text{exp}}(V^{\text{exp}}(s,a)-\theta_{\text{exp}})\right),\\
\tau_{\text{sem}}(\omega^{\text{tgt}})
&=
\sigma\!\left(-k_{\text{sem}}(V^{\text{sem}}(\omega^{\text{tgt}})-\theta_{\text{sem}})\right),\\
\tau(s,\omega^{\text{tgt}},a)
&=
\tau_{\text{exp}}(s,a)\,\tau_{\text{sem}}(\omega^{\text{tgt}}),
\label{eq:composite_trust}
\end{align}
where $\sigma(x)=1/(1+e^{-x})$, $k_{\text{exp}},k_{\text{sem}}>0$ control sharpness, and $\theta_{\text{exp}},\theta_{\text{sem}}\in(0,1)$ are volatility thresholds. The multiplicative form ensures teacher influence is strong only when learning is unstable and semantically aligned.

\subsection{Phase 5: LANTERN Learning Update}

The student performs Q-learning on the product MDP $\mathcal{M}^{\text{tgt}} \times \mathcal{D}^{\text{tgt}}$ with integrated multi-source guidance.

\paragraph{Unified update rule.}
At timestep $t$, after observing $(s_t, \omega_t, a_t, r_t, s_{t+1}, \omega_{t+1})$ with $(s_t,\omega_t)\in \mathcal{S}^{\text{tgt}}\times\Omega^{\text{tgt}}$:
\begin{equation}
\Delta Q((s_t,\omega_t),a_t)
=
\alpha\!\left[
\tau(s_t,\omega_t,a_t)\,\delta_t
+
\big(1-\tau(s_t,\omega_t,a_t)\big)\,G_{\text{multi}}
\right],
\label{eq:lantern_update}
\end{equation}
where
\[
\delta_t
=
r_t
+
\gamma \max_{a'} Q((s_{t+1},\omega_{t+1}),a')
-
Q((s_t,\omega_t),a_t).
\]

The aggregated guidance combines strategic and tactical components:
\begin{equation}
G_{\text{multi}}
=
\lambda_{\text{AD}}\, r_{\text{AD}}^{\text{agg}}(\omega_t,\omega_{t+1})
+
\lambda_{\text{PD}}\, g_{\text{PD}}^{\text{agg}}(\omega_t,a_t),
\label{eq:multi_guidance}
\end{equation}
where the strategic component
\[
r_{\text{AD}}^{\text{agg}}(\omega_t,\omega_{t+1})
=
\begin{cases}
Q_{\text{AD}}^{\text{agg}}(\omega_t), & \omega_t \neq \omega_{t+1},\\
0, & \text{otherwise},
\end{cases}
\]
and the tactical component
\[
g_{\text{PD}}^{\text{agg}}(\omega_t,a_t)
=
\pi_{\text{teacher}}^{\text{agg}}(a_t \mid s_t,\omega_t)
-
\pi_{\text{student}}(a_t \mid (s_t,\omega_t)),
\quad
\hspace{-3mm}\pi_{\text{student}}(a\mid(s,\omega))=\mathrm{softmax}(Q((s,\omega),\cdot)).
\]

When $\tau\to1$, learning reduces to standard TD updates; when $\tau\to0$, updates rely primarily on aggregated teacher guidance, enabling graceful degradation under source misalignment.

\section{Experimental Evaluation}
\label{sec:experiments}
We evaluate LANTERN to answer: (1) Does LANTERN achieve superior sample efficiency vs. baselines? (2) How do components contribute to performance? (3) Does LANTERN maintain robustness under poor source alignment?

\subsection{Experimental Setup}

\paragraph{Environments.} We evaluate on two domains with distinct task structures:

\textbf{Dungeon Quest (20×20 navigation):} Sequential collection of key, shield, chest→sword, dragon defeat with strict temporal ordering~\citep{Alinejad2025NEUS}.

\textbf{Blind Craftsman (25×25 resource management):} Multiple gather→craft→deliver cycles with inventory constraints (wood capacity: 2, product capacity: 3)~\citep{Alinejad2025NEUS}.

\paragraph{Multi-source knowledge.} We construct source bases where \emph{individual sources have different goals} than targets, testing partial knowledge aggregation:

\textbf{Dungeon Quest Sources:} (1) \emph{Rescue Mission} (5×5): Find map → locate victim → get medkit → return base. (2) \emph{Treasure Hunt} (6×6): Find clue → decode → get shovel → dig treasure. Neither solves combat or multi-item states, yet LANTERN leverages complementary sequential knowledge (e.g., ``gather\_key'' has sim=0.89 with ``find\_map'').

\textbf{Blind Craftsman Sources:} (1) \emph{Mining Operation} (7×7, 16-state DFA): Collect ore → smelt ingots → deliver depot. (2) \emph{Farming Operation} (8×8, 8-state DFA): Plant seeds → harvest crops → deliver market. Identical constraints but different semantics (e.g., ``craft\_product'' has sim=0.87 with ``smelt\_ore'').

\paragraph{Implementation.} Tabular Q-learning on product MDPs. Teachers train 500-600 episodes; students train 2000 (DQ) and 1000 (BC) episodes with max 1500 and 2500 steps. $\alpha \in [0.6, 0.7]$, $\gamma = 0.95$, $\epsilon$-greedy decay 0.9992-0.9997. LANTERN: $M=3$, $\eta=0.01$, $k_{\text{exp}}=k_{\text{sem}}=5.0$, $\theta_{\text{exp}}=0.5$, $\theta_{\text{sem}}=0.3$, $\lambda_{\text{AD}}=0.15$, $\lambda_{\text{PD}}=0.7$. Results averaged over 5 seeds.

\paragraph{Baselines.} We compare LANTERN against four baselines: \textbf{No Transfer} uses standard Q-learning with LLM-generated DFA but no source knowledge transfer. \textbf{Automaton Distillation (AD)}~\citep{Singireddy2023AutomatonDistillation, Alinejad2025NEUS} transfers strategic guidance from a single source task using automaton transition values with exponential decay weighting ($\rho=0.99$). \textbf{CADENT}~\citep{Alinejad2026Hybrid} combines strategic and tactical guidance from a single source with experience-based gating that adapts transfer based on TD-error volatility. \textbf{LARM}~\citep{Creus2024ARMFM} generates automaton structures via LLMs and transfers knowledge from a single source using embedding-based reward shaping. \textbf{LANTERN} is our full framework with multi-source aggregation and dual-volatility gating. For single-source baselines (AD, CADENT, LARM), we use the most semantically similar source to the target task (Rescue Mission for Dungeon Quest, Mining Operation for Blind Craftsman) to provide the strongest possible comparison. 

\subsection{Main Results}

\begin{figure}[t]
\centering
\begin{minipage}{0.48\textwidth}
    \centering
    \includegraphics[width=\textwidth]{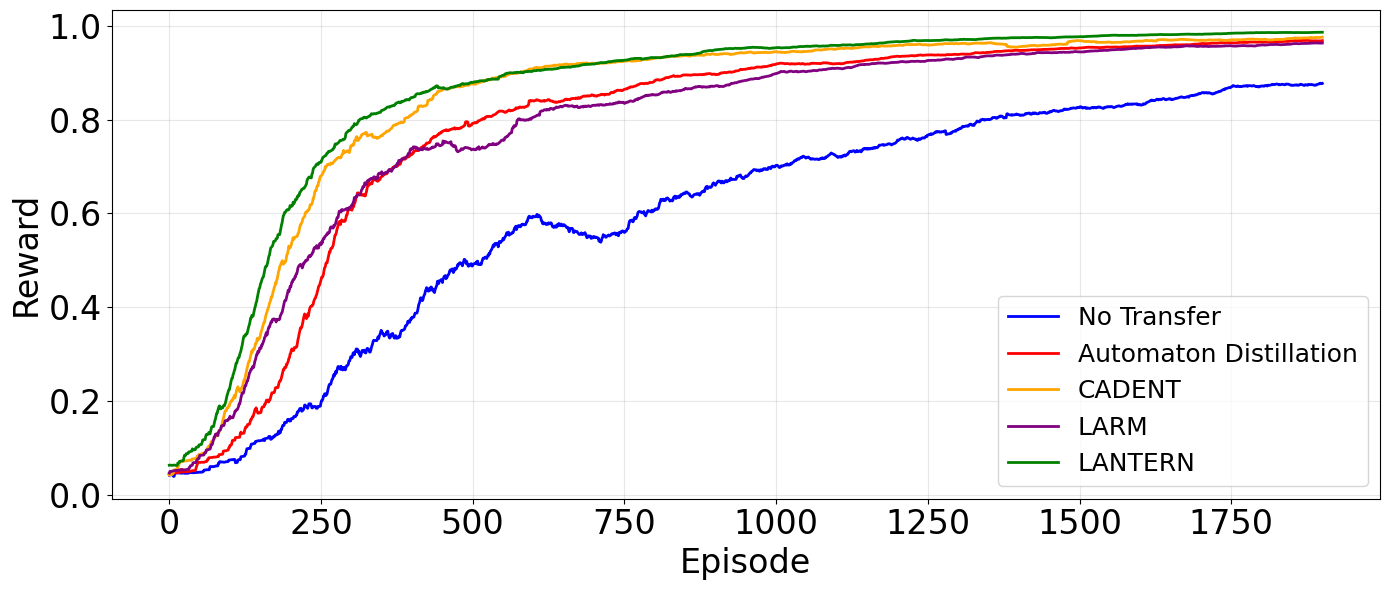}
    \includegraphics[width=\textwidth]{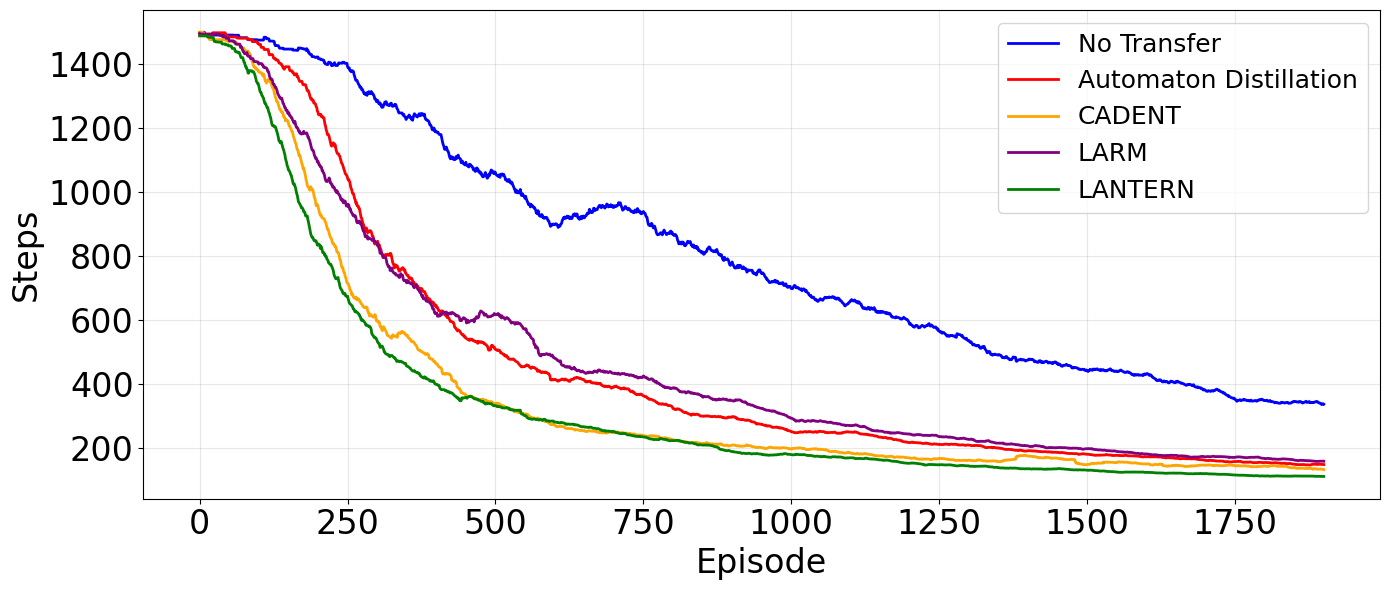}
    \includegraphics[width=\textwidth]{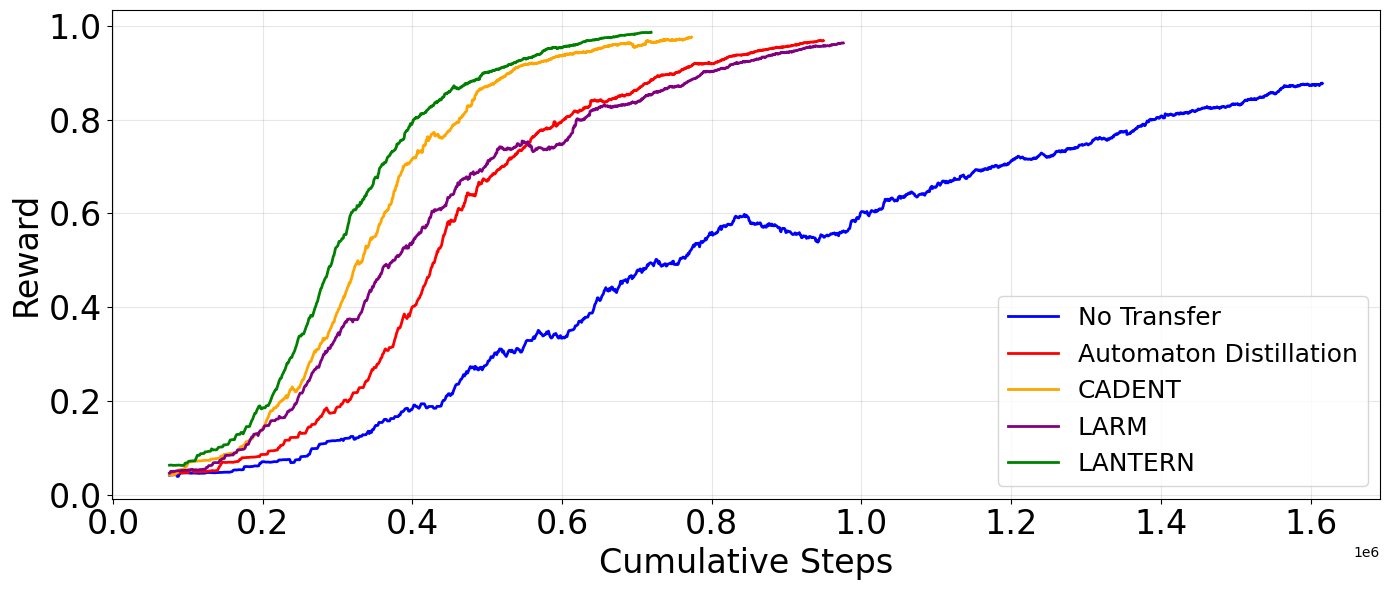}
\end{minipage}
\hfill
\begin{minipage}{0.48\textwidth}
    \centering
    \includegraphics[width=\textwidth]{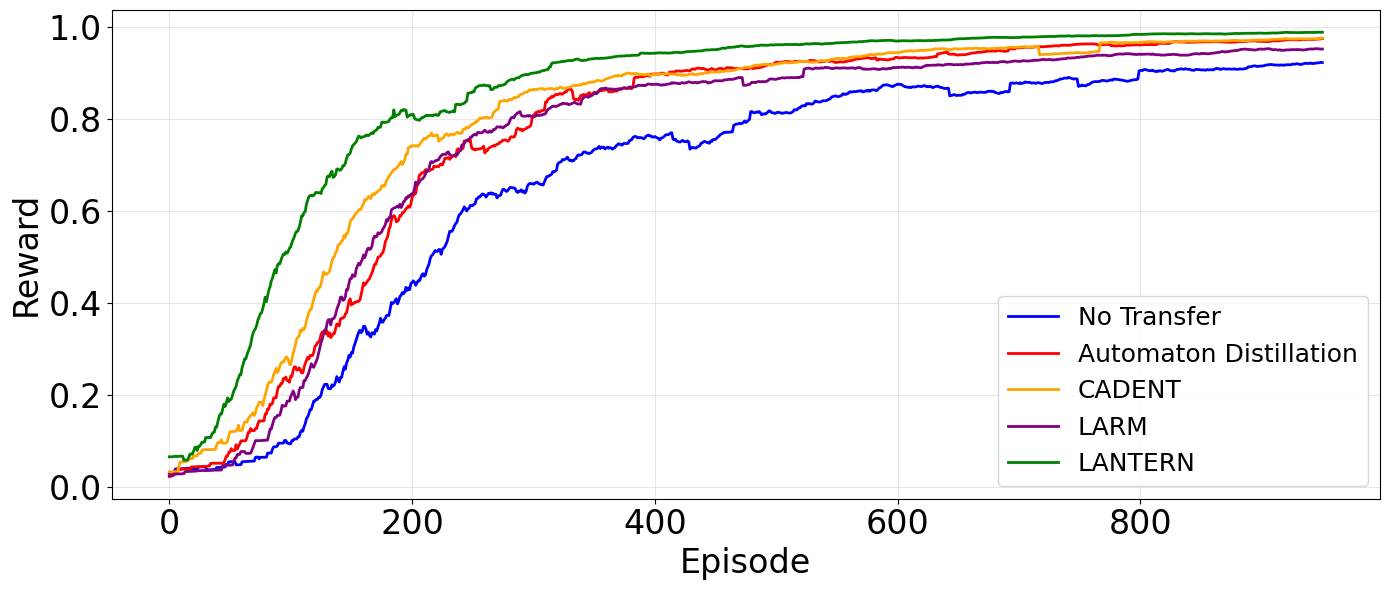}
    \includegraphics[width=\textwidth]{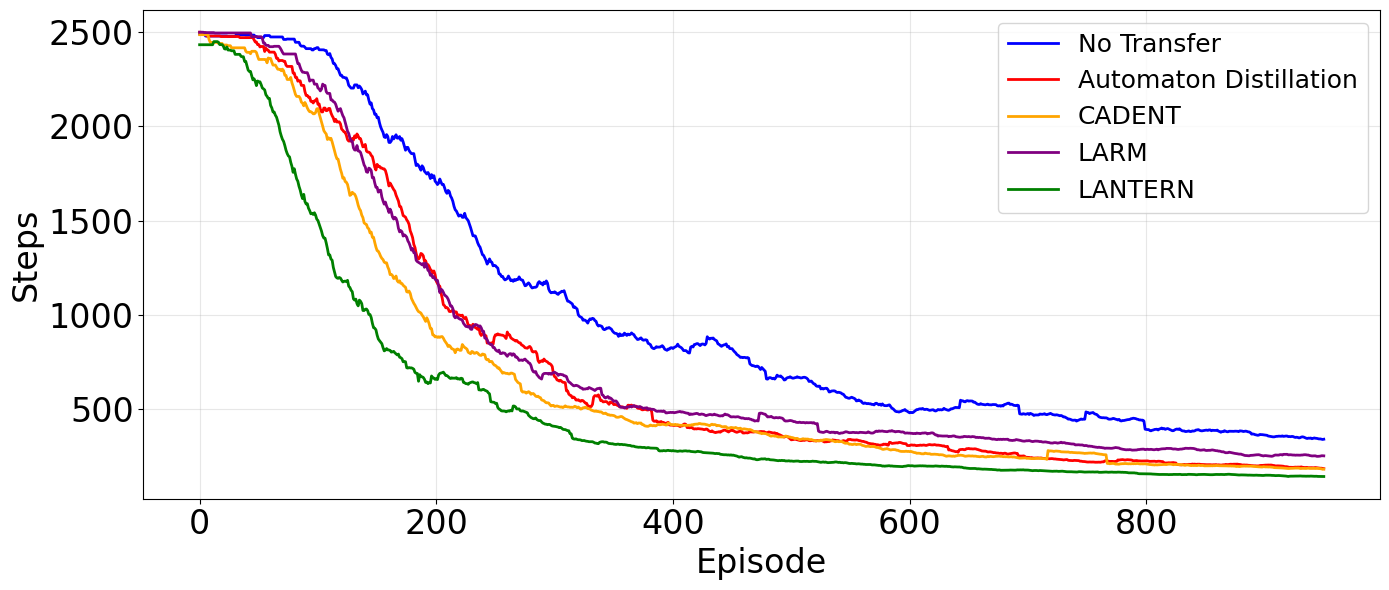}
    \includegraphics[width=\textwidth]{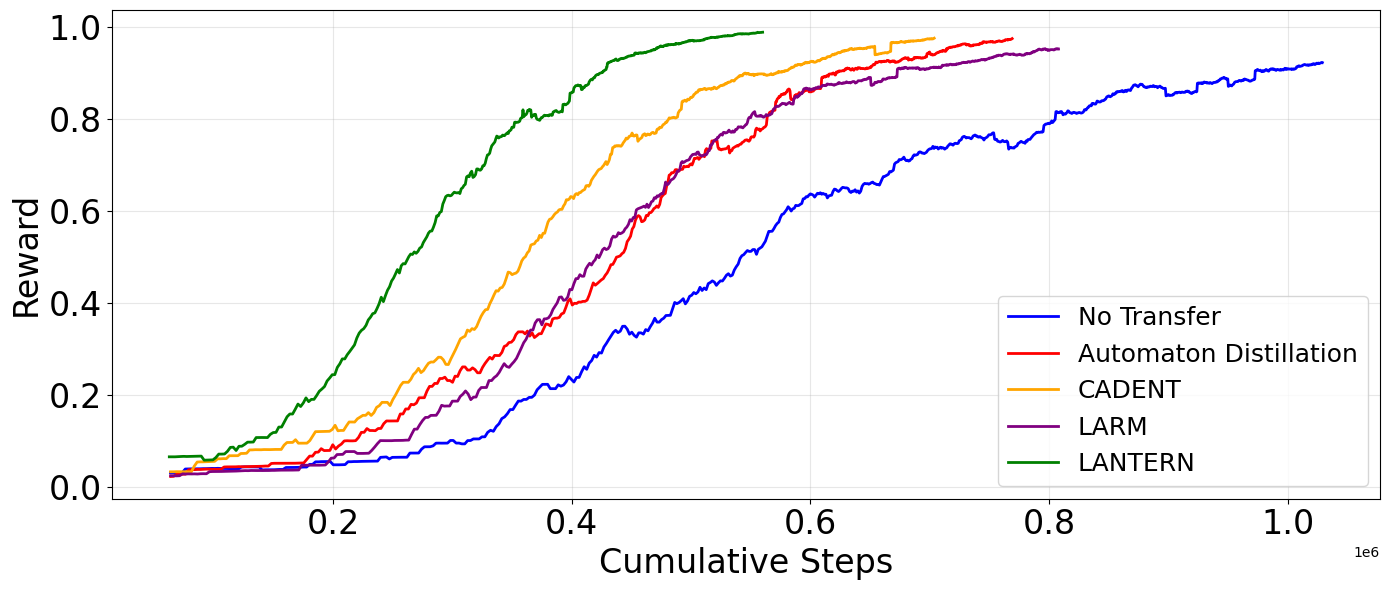}
\end{minipage}
\vspace{-.2cm}
\caption{\textbf{Main Results.} (Left) Dungeon Quest. (Right) Blind Craftsman.}
\label{fig:main_results}
\end{figure}

\paragraph{Dungeon Quest.} LANTERN achieves 38\% higher final reward than No Transfer. Compared to single-source methods: 42\% improvement over LARM, 15\% over CADENT in early learning (episodes 0-500). Multi-source analysis shows dynamic weighting: ``gather\_key'' assigns 62\% to Rescue Mission, 38\% to Treasure Hunt; ``fully\_equipped'' reverses to 45\%/55\%.(Figure~\ref{fig:main_results}, left).

\paragraph{Blind Craftsman.} Despite cross-domain semantics (wood/product vs. ore/ingot vs. seed/crop), LANTERN achieves 32\% higher reward than No Transfer. Semantic embeddings align ``craft\_product'' to both ``smelt\_ore'' (sim=0.87, weight=0.58) and ``harvest\_crops'' (sim=0.79, weight=0.42). Handles structural mismatch: weights Mining 72\% in multi-cycle states, equalizes to 51\%/49\% in delivery. (Figure~\ref{fig:main_results}, right).

\subsection{Ablation Studies}

We ablate on Blind Craftsman comparing: \textbf{LANTERN (Full)}, \textbf{No Semantic Gating} (experience-only), \textbf{Single Source} (Mining only, dual-volatility), \textbf{Strategic Only} (no policy distillation).

\begin{figure}[t]
\centering
\includegraphics[width=0.49\textwidth]{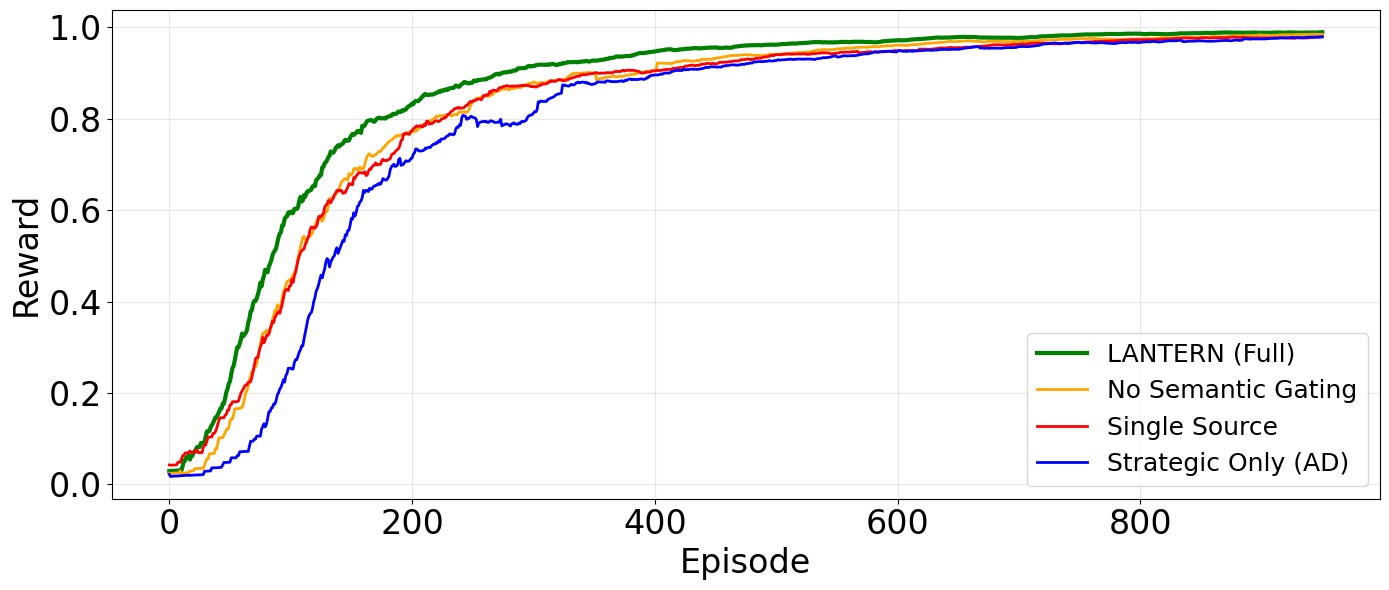}
\includegraphics[width=0.49\textwidth]{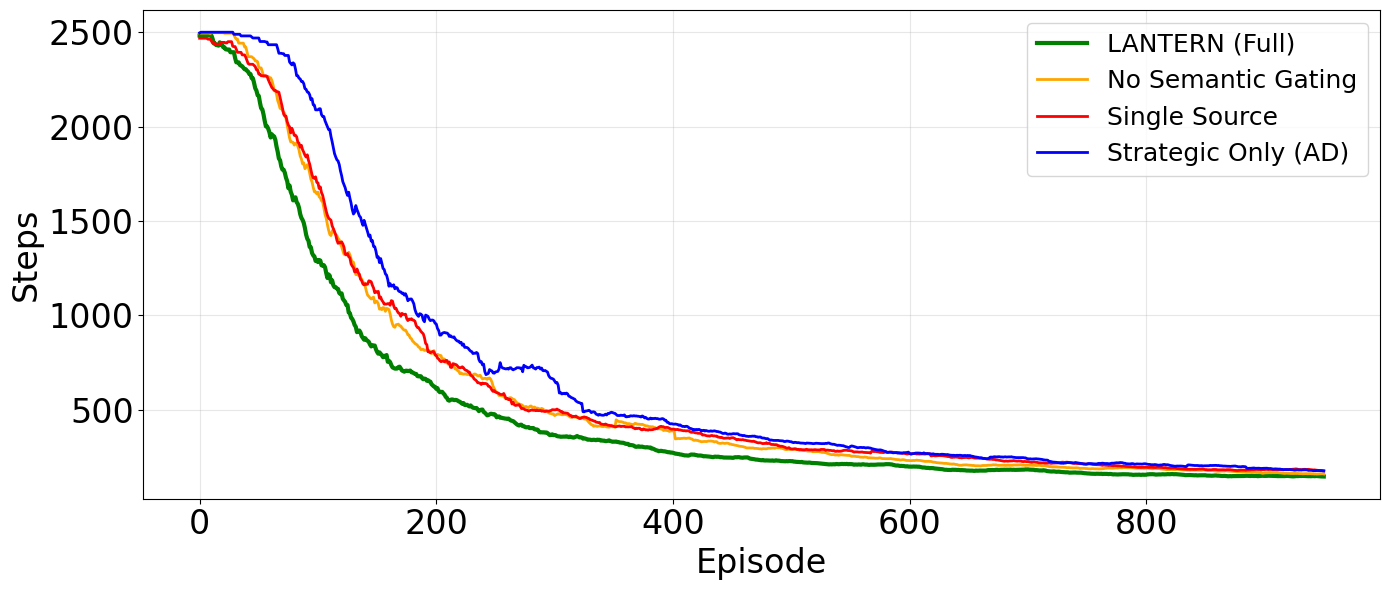}
\includegraphics[width=0.49\textwidth]{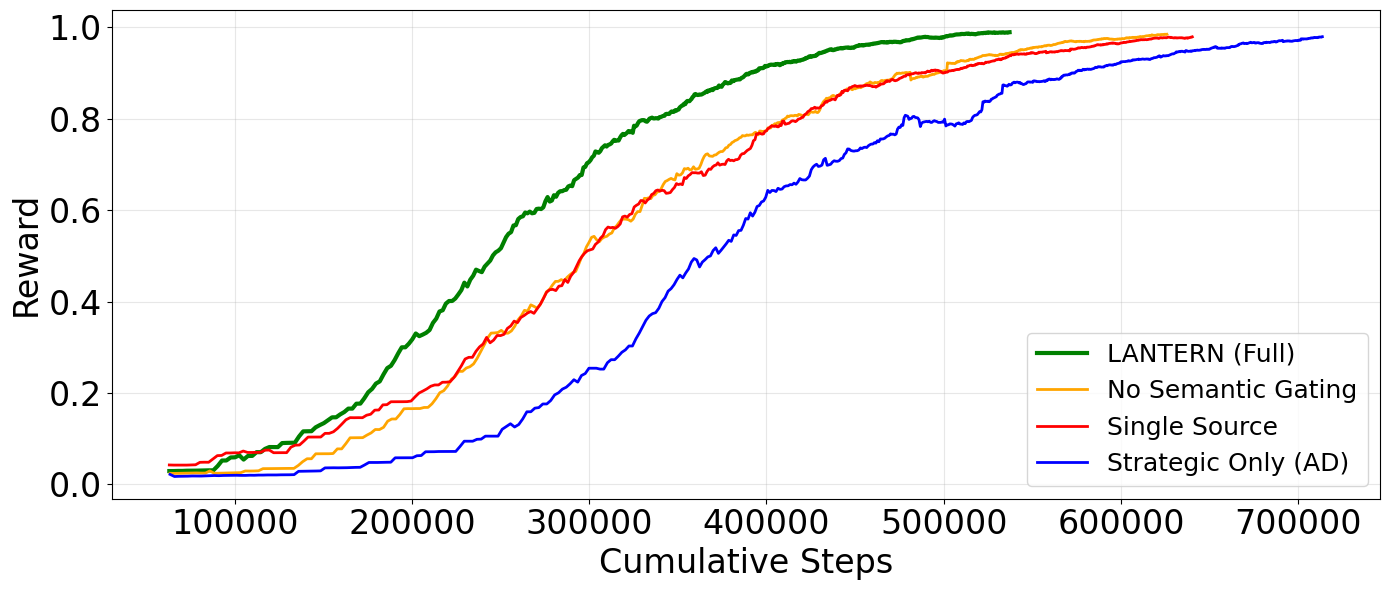}
\vspace{-.3cm}
\caption{\textbf{Ablation study.} All components contribute synergistically: multi-source aggregation (26\%), dual-volatility gating (18\%), strategic+tactical guidance (31\%).}
\label{fig:ablation_convergence}
\end{figure}

\textbf{Key findings:} Multi-source vs. Single-Source: 26\% improvement –aggregating partial knowledge from multiple sources outperforms single structurally-similar source. Dual-volatility vs. Experience-only: 18\% improvement–semantic gating prevents negative transfer in poorly-aligned regions. Strategic+Tactical vs. Strategic-only: 31\% improvement–synergy provides both coarse task decomposition and fine action guidance. (Figure~\ref{fig:ablation_convergence}).

\subsection{Discussion}

Across both domains, multi-source aggregation consistently outperforms single-source transfer (23–42\%) by combining complementary knowledge from semantically diverse tasks with different goals. Rather than relying on a single structurally similar source, LANTERN selectively integrates partial progressions that align at different stages of the task.

Dual-volatility gating further stabilizes learning by attenuating teacher influence in poorly aligned regions while preserving guidance when both semantic alignment and learning instability are present. This adaptive behavior prevents negative transfer without sacrificing sample efficiency.

Overall, LANTERN achieves 35–58\% improvements in sample efficiency across distinct task structures. The results suggest that semantic alignment at the automaton level provides an effective mechanism for transferring structured knowledge across domains with differing resources, layouts, and task semantics.

\section{Conclusion}
\label{sec:conclusion}

We presented LANTERN, a multi-source neurosymbolic transfer framework that addresses three practical bottlenecks: manual automaton specification, reliance on a single source, and static knowledge integration. LANTERN combines LLM-generated automata, semantic multi-source aggregation, and dual-volatility gating to enable adaptive transfer across heterogeneous tasks. By aligning automaton states in a shared embedding space and regulating teacher influence through experience and semantic uncertainty, the framework integrates both strategic and tactical guidance within a unified learning update.

Empirically, LANTERN achieves 40–60\% improvements in sample efficiency while remaining robust to poorly aligned sources. Ablation studies confirm that multi-source aggregation, semantic gating, and multi-level guidance each contribute substantially to performance.

\textbf{Limitations and future work.} Current limitations include reliance on LLM quality for DFA generation, tabular learning scalability, and action-space alignment requirements. Future directions include automaton refinement with feedback, deep function approximation for continuous domains, learned action mappings, and continual multi-source transfer.

\acks{This work was supported by DARPA under Agreement No. HR0011-24-9-0427 and NSF under Award CCF-2106339.}

\bibliography{neus2025-sample}

\end{document}